%% file: root.tex
\documentclass[a4paper, 10pt, conference]{ieeeconf}      

\IEEEoverridecommandlockouts                              

\input{preamble/preamble}

\title{\LARGE \bf
ParCo-SDF: Learning Prior-Free Partial-to-Complete \\
Signed Distance Fields of Deformable Objects
}

\author{Deokmin Hwang$^{}$, Minseok Song, and Daehyung Park$^{\dag}$
\thanks{All authors are with the School of Computing, Korea Advanced Institute of Science and Technology, Korea
        {\tt\small \{gsh04089, hjmngb, daehyung\}@kaist.ac.kr}}%
\thanks{$^{\dag}$D. Park is the corresponding author.
        }%
}

\begin{document}

\maketitle
\thispagestyle{empty}
\pagestyle{empty}

\begin{abstract}
This study addresses the partial-to-complete geometry reconstruction of deformable objects (DOs) from point-cloud observations toward precise DO manipulation. Recent DO reconstruction approaches often adopt implicit neural representations (INRs) to model continuous surfaces as well as capture structural variability. However, these methods typically rely on object-specific shape priors that improve training stability and limit generalization. To figure it out, we introduce ParCo-SDF, a two-stage partial-to-complete signed distance field (SDF) reconstruction framework consisting of temporal geometry encoding followed by FiLM-conditioned SDF prediction. The temporal encoder captures structural similarity across DO sequence, enabling prior-free stable training. FiLM-based conditioning preserves reconstruction expressivity while reducing network complexity. We evaluate the proposed method against a state-of-the-art DO surface reconstruction baseline on a rubber band manipulation dataset, demonstrating robust and high-fidelity reconstruction under severe occlusions.

\end{abstract}

\input{sec/1_intro}
\input{sec/4_method}

\input{sec/5_experiment}
\input{sec/7_dicussion}




\section*{ACKNOWLEDGMENT}
This research was supported by the IITP (Institute of Information \& Communications Technology Planning \& Evaluation)-ITRC (Information Technology Research Center) grant funded by the Korea government (Ministry of Science and ICT, MSIT) (IITP-2026-2024-00437102), the National Research Foundation of Korea (NRF) grant funded by the MSIT (RS-2026-25482412), Artificial intelligence industrial convergence cluster development project funded by the MSIT \& Gwangju Metropolitan City, and IITP grants funded by the MSIT (No. RS-2024-00509279, Global AI Frontier Lab, and RS-2022-II220311).



\bibliographystyle{IEEEtran}
\bibliography{references}

\end{document}

%% file: preamble/preamble.tex
\usepackage{verbatim}
\usepackage{amssymb}
\usepackage{amsmath}
\usepackage{amsfonts} 
\usepackage{kotex}

\usepackage{amsthm} 
\usepackage{graphicx} 
\usepackage{mathtools} 
\usepackage{enumitem}
\usepackage{xspace}
\usepackage[ruled,vlined]{algorithm2e}
\makeatletter
\let\NAT@parse\undefined
\makeatother
\usepackage{hyperref}
\usepackage[dvipsnames]{xcolor}
\hypersetup{
    colorlinks=true,
    linkcolor=magenta,
    citecolor=blue,
    }
\usepackage[all]{hypcap}   
\usepackage{stfloats}
\usepackage{siunitx}

\input{preamble/math}

%% file: preamble/math.tex
\newcommand{\bb}{\mathbf{b}}
\newcommand{\bc}{\mathbf{c}}

\newcommand{\bn}{\mathbf{n}}

\newcommand{\bp}{\mathbf{p}}

\newcommand{\bv}{\mathbf{v}}

\newcommand{\bx}{\mathbf{x}}

\newcommand{\bz}{\mathbf{z}}

\newcommand{\bK}{\mathbf{K}}

\newcommand{\bQ}{\mathbf{Q}}

\newcommand{\bV}{\mathbf{V}}
\newcommand{\bW}{\mathbf{W}}

\newcommand{\bZ}{\mathbf{Z}}

\newcommand{\calE}{{\mathcal{E}}}
\newcommand{\calF}{{\mathcal{F}}}
\newcommand{\calG}{{\mathcal{G}}}

\newcommand{\calL}{{\mathcal{L}}}
\newcommand{\calM}{{\mathcal{M}}}
\newcommand{\calN}{{\mathcal{N}}}

\newcommand{\calP}{{\mathcal{P}}}
\newcommand{\calQ}{{\mathcal{Q}}}

\newcommand{\calU}{{\mathcal{U}}}
\newcommand{\calV}{{\mathcal{V}}}

\newcommand{\bbR}{\mathbb{R}}
\newcommand{\bbS}{\mathbb{S}}

\newcommand{\bbeta}{{\boldsymbol{\beta}}}

\newcommand{\btheta}{{\boldsymbol{\theta}}}

\def\[#1\]{\begin{equation}\begin{aligned}#1\end{aligned}\end{equation}}

\makeatother

%% file: sec/1_intro.tex
\section{Introduction}
Deformable object manipulation (DOM) is a fundamental yet challenging problem in robotics~\cite{hang2021modeling, khazatsky2024droid, oh2026estimation}.
Unlike rigid objects, deformable objects (DOs) exhibit infinite degrees of freedom (DoF), undergoing continuous geometric and topological changes during interaction.
These properties frequently cause severe self-occlusions and robot-induced occlusions, obscuring critical state information required for manipulation.
The occluded structures and their surface representations play a critical role in downstream manipulation and control tasks~\cite{kim2022graphdistnet,oh2024dispo,kim2025visuo} by characterizing deformation, contact interactions, and physical responses~\cite{barbany2024deformable, hanxiao2025phystwin}.
Consequently, recovering complete object geometry from partial sensory observations has become a central problem in deformable object manipulation.

Early approaches rely predominantly on explicit representations, such as point clouds and meshes~\cite{deng2022survey,huang2022mesh}.
Point cloud representations suffer from finite sampling and undesired point distributions. Further, the absence of explicit structural connectivity makes them vulnerable to severe overlap and occlusion~\cite{li2018learning}.
Mesh-based representations explicitly encode surface topology; however, capturing fine-grained deformations requires dense discretization, which introduces computational cost~\cite{yusuke2023deformable}.
Recent advances in implicit neural representations (INRs) offer a compelling alternative by modeling geometry as continuous functions over space, allowing resolution-independent and topology-agnostic surface reconstruction~\cite{ruyi2023endosurf,prokudin2023dynamic}. 
In particular, signed distance fields (SDFs) encode not only surface geometry but also surface normals via spatial gradients. 
This property allows SDF-based models to represent severe deformations with self-contact, making them well-suited for DO representation~\cite{ruyi2023endosurf}.
Despite their effectiveness, INR-based approaches often focus on memorizing a single DO, thereby limiting their ability to generalize to unseen deformation scenarios.

\begin{figure}[t]
    \centerline{\includegraphics[width=\linewidth]{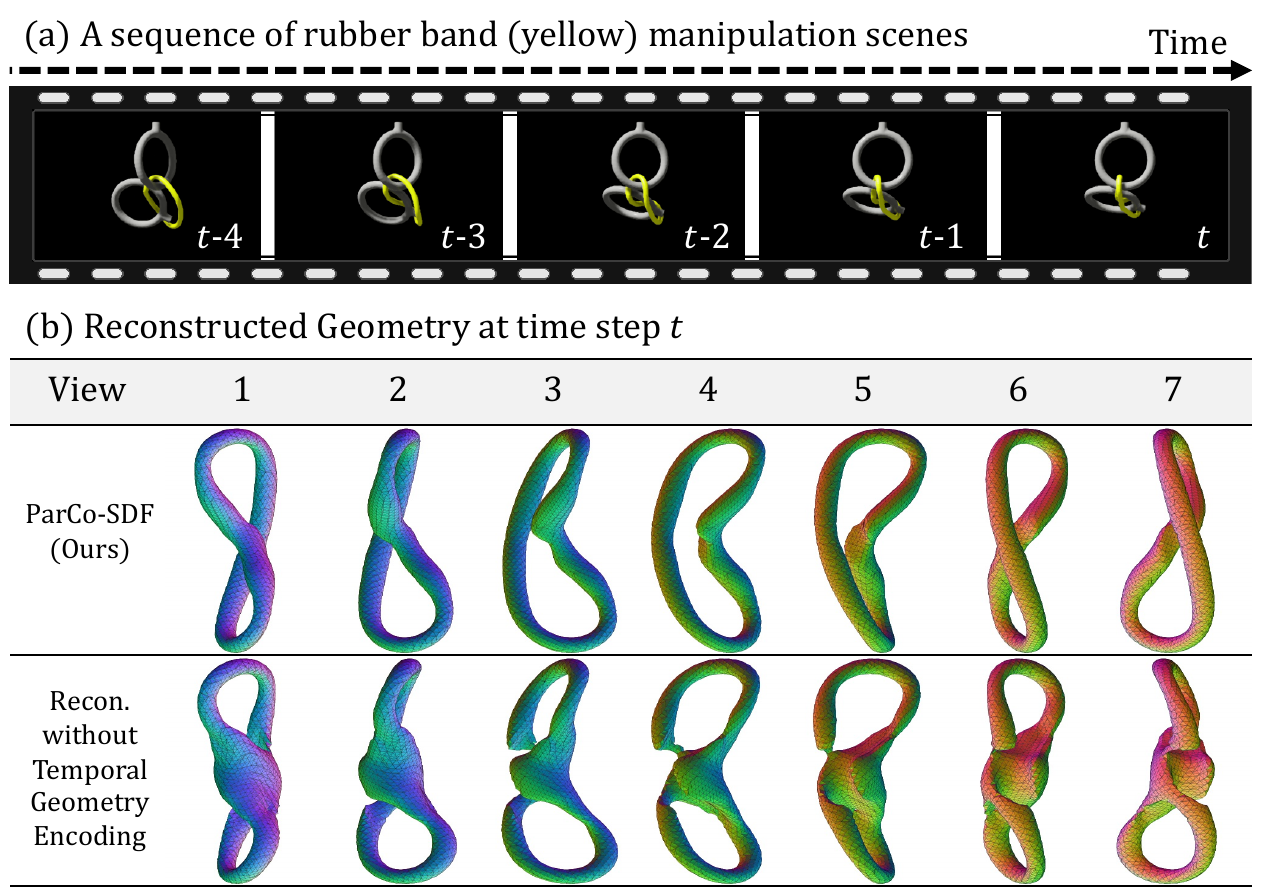}}
    \caption{Comparison of partial-to-complete reconstruction performances under severe deformation and occlusion of the rubber band (yellow).    
(a) A sequence of rubber band manipulation scenes illustrating the acquisition of partial observations (i.e., point clouds) used for temporal geometry encoding. Note that the last scene at time step $t$ exhibits severe occlusion, resulting in highly partial observations.
(b) Reconstructed geometry at time step $t$ with and without temporal geometry encoding, visualized from seven viewpoints.
The proposed method, ParCo-SDF, infers the complete geometry from temporal observations, enabling occlusion-robust reconstruction.
}
\label{fig:recon}
\end{figure}

To extend INRs beyond single-instance reconstruction, prior works introduce conditioning mechanisms that inject instance-specific information into a shared implicit network~\cite{saito2023neural}.
A common strategy, \textit{input concatenation}, augments implicit network query by concatenating an instance-specific latent code~\cite{park2019deepsdf}. However, as the number of instances increases, the limited scalability of the shared implicit network constraints its representational capacity, thereby restricting the expressiveness across diverse instances. 
Hypernetwork-based methods alleviate this limitation by predicting implicit network parameters from latent codes~\cite{sitzmann2020siren,wi2022virdo}, but introduce significant computational overhead and training instability due to the quadratic growth of predicted parameters with network width~\cite{alaluf2022hyperstyle}.
This highlights the need for a conditioning mechanism that jointly models continuously evolving deformation states while maintaining both expressive capacity and stable training.

Beyond representation and conditioning challenges, reconstructing complete object geometry from partial observations is inherently ill-posed~\cite{hartley2003multiple, dai2017shape}.
Researchers often rely on shape priors, such as predefined templates or geometric symmetry, to achieve occlusion-robust reconstruction~\cite{prokudin2023dynamic, kohei20243d, song2025implicit}.
However, such assumptions restrict generalization to unforeseen structures or deformations that do not conform to these priors.
Alternatively, multi-view approaches mitigate single-frame ambiguities by aggregating observations from multiple viewpoints~\cite{wang2021neus}.
However, under large deformations such as twisting, object parts may collapse or self-occlude, resulting in the loss of reliable geometric cues.

In this context, we propose ParCo-SDF, a partial-to-complete implicit SDF reconstruction framework leveraging temporal geometry encoding. ParCo-SDF consists of two stages: (1) temporal geometry encoding and (2) modulation-based SDF prediction.
The temporal geometry encoder aggregates a sequence of partial observations into a latent code via self-attention~\cite{vaswani2017attention}, capturing structural information without imposing explicit shape priors (see Fig.~\ref{fig:recon}).
Conditioned on the code, the modulation-based SDF predictor then reconstructs an implicit SDF network to complete the full DO geometry via feature-wise linear modulation (FiLM)~\cite{perez2018film}. This modulation structure helps increasing the implicit network size while preserving training stability.

We evaluate ParCo-SDF on a rubber band manipulation dataset. The results show that our method consistently outperforms a baseline in geometric fidelity and robustness to previously unseen deformation states, demonstrating occlusion-robust reconstruction across diverse structures and deformations.

%% file: sec/4_method.tex
\begin{figure*}[t]
    \centerline{\includegraphics[width=\linewidth]{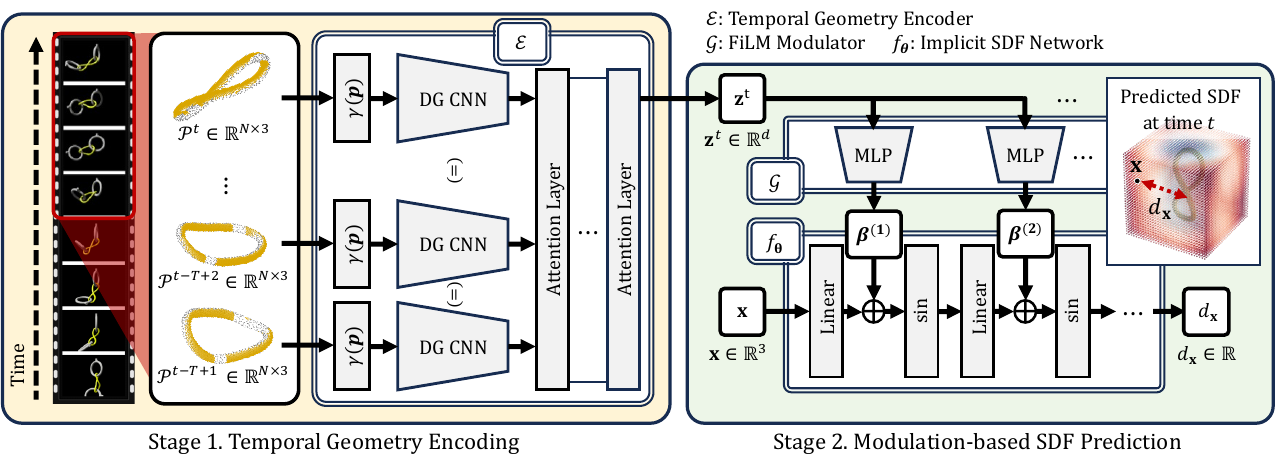}}
    \caption{
        Overview architecture of ParCo-SDF for reconstructing signed distance functions (SDFs) of deformable objects from a sliding window of partial point-cloud observations $\{\calP^{t-T+1},\ldots,\calP^t\}$.
        The temporal geometry encoder $\calE$ aggregates a sequence of partial point clouds through Fourier-feature mapping $\gamma$, DGCNN-based local geometry encoding, and attention-based temporal aggregation to produce a latent code $\mathbf{z}^t$. 
        Given $\bz^t$, a modulator $\calG$ predicts layer-wise shift parameters, which modulate a shared implicit network $f_\btheta$ that returns the signed distance $d_\bx$ at a queried spatial coordinate $\bx$.
        Here, $\oplus$ denotes the element-wise sum.
    } \label{fig:overview}
\end{figure*}

\section{Method}
We address the partial-to-complete geometry reconstruction of DOs from temporally ordered partial point-cloud observations acquired during manipulation.
At each time step $t$, a fixed external depth sensor captures a partial point cloud $\calP^t \subset \bbR^3$ of the DO.
Note that due to self-occlusion and robot-induced occlusion, each observation alone lacks sufficient geometric cues to reconstruct the complete structure of the deformation state.

We represent the complete object geometry as the zero level-set of an implicit signed distance function defined over a bounded spatial domain $\calQ \subset \bbR^3$.
Given a query point $\bx \in \calQ$, an implicit neural network 
$f_{\btheta} : \calQ \rightarrow \bbR$, parameterized by $\btheta$, 
predicts a signed distance value to the object surface,
where positive and negative values indicate the exterior and interior of the object, respectively.
To model the evolving geometry with a single shared implicit network,
we condition the network on a latent code extracted from a sliding window of temporally ordered partial observations,
$\{\calP^{t-T+1}, \ldots, \calP^t\}$,
where $T$ denotes the window length.

As illustrated in Fig.~\ref{fig:overview}, the overall architecture consists of two stages:
(1) temporal geometry encoding (see Sec.~\ref{subsec:TGE}) and 
(2) modulation-based SDF prediction (see Sec.~\ref{subsec:MSR}).
The encoder $\calE$ maps the observation window to a latent code $\bz^t \in \bbR^d$, 
where $d$ denotes the latent dimensionality.
Conditioned on $\bz^t$, the implicit SDF reconstructor $f_{\btheta}(\bx; \bz^t)$ predicts the signed distance at $\bx$.
In the following subsections, we describe each component and the training procedure in detail.

\subsection{Temporal Geometry Encoder}
\label{subsec:TGE}

We extract a deformation-aware latent code from a sliding window of partial point clouds,
$\{\calP^{t-T+1}, \ldots, \calP^t\}$.
Instead of directly aggregating raw point sets across time—which is sensitive to irregular sampling and lacks correspondence under large deformations—our encoder transforms each observation into a compact geometry code and performs aggregation in the latent space.
The encoder consists of three stages:
(1) point-level feature lifting,
(2) local geometric encoding per observation, and
(3) temporal aggregation of observation-level codes.
Each stage addresses a distinct challenge in deformation-aware geometry inference.

\subsubsection{Point-level feature lifting}

We first enhance the spatial expressiveness of raw coordinates by lifting each point into a higher-dimensional feature space.
Given a point $\bp \in \calP^i$, where $i \in \{t-T+1,\ldots,t\}$,
we apply Fourier-feature mapping~\cite{tancik2020fourier} to obtain
\begin{equation}
    \gamma(\bp) = \left[\cos(2\pi \mathbf{B}\bp),\;\sin(2\pi \mathbf{B}\bp)\right],
\end{equation}
where $\mathbf{B} \in \bbR^{m \times 3}$ denotes the Fourier basis matrix and $m$ is the number of bases.
Tancik \textit{et al}.~\cite{tancik2020fourier} originally sample $\mathbf{B}$ from a Gaussian distribution and keep it fixed during training.
In contrast, we treat $\mathbf{B}$ as a learnable parameter so that the encoder can adapt its spectral encoding to diverse deformation states.
In this work, we set $m=32$.

\subsubsection{Local geometric encoding per observation}

Next, we encode each lifted observation into a compact geometry code.
Given a lifted observation
$\tilde{\calP}^i = \{\gamma(\bp) \mid \bp \in \calP^i\}$,
we employ a dynamic graph convolutional neural network (DGCNN)~\cite{wang2019dynamic} to produce an observation-level latent code $\tilde{\bz}^i \in \bbR^{d}$.
DGCNN constructs a $k$-nearest neighbor graph and updates point-wise features via edge-based convolutions, explicitly modeling local spatial relationships.
This stage yields a permutation-invariant and geometry-aware representation that captures partial structure.

\subsubsection{Temporal aggregation of observation-level codes}
Finally, we aggregate the sequence of observation-level codes to infer the deformation state at the time step $t$.
Although each $\tilde{\bz}^i$ encodes geometry-aware information, it reflects only a partially observed surface.
To integrate complementary geometric evidence across time,
we stack the codes $\bZ = [\tilde{\bz}^{t-T+1}, \ldots, \tilde{\bz}^t]$ and apply self-attention layers~\cite{vaswani2017attention}.
At each attention layer, we compute
$\bQ = \bZ \bW_Q$, $\bK = \bZ \bW_K$, and $\bV = \bZ \bW_V$,
where $\bW_Q, \bW_K, \bW_V \in \bbR^{d \times d}$ are learnable parameters.
We update the representations via scaled dot-product attention:
\begin{equation}
\mathrm{Attn}(\bQ, \bK, \bV)
=
\mathrm{softmax}\!\left(\frac{\bQ \bK^\top}{\sqrt{d}}\right)\bV.
\end{equation}
We take the output corresponding to the final time step as the temporally aggregated latent code $\bz^t \in \bbR^d$,
which serves as a deformation-aware geometry representation for conditioning the implicit SDF reconstructor.

\subsection{FiLM-Conditioned Surface Reconstructor}
\label{subsec:MSR}
To incorporate instance-specific geometric information, we condition the implicit neural network using FiLM-based modulation~\cite{perez2018film}.
We adopt a SIREN architecture~\cite{sitzmann2020siren} as the backbone of the implicit network to enable accurate modeling of high-frequency geometric details~\cite{sitzmann2020siren}.
In contrast to the original SIREN formulation, our SIREN improves optimization stability across diverse deformed geometries of the DO 
by applying the layer-wise scaling factor $\omega_0$ only to the first hidden layer, which otherwise amplifies gradients during training~\cite{de2022simple}.

We design shift-only modulation, which avoids amplifying sinusoidal pre-activations, to improve optimization stability~\cite{sitzmann2020siren}.
We employ a set of modulation networks $\{\calG^{(\ell)}\}_{\ell=1}^{L-1}$ that map the latent representation $\bz^t$ to layer-wise shift parameters $\bbeta^{(\ell)}$, where $L$ denotes the number of layers in the implicit network.
We define the $\ell$-th layer output of FiLM-conditioned implicit network $f^{(\ell)}_\btheta$ recursively as
\begin{align}
    \label{eq:siren}
    \bbeta^{(\ell)} &= \calG^{(\ell)}(\bz^t), \quad \ell\in\{1,\ldots,L-1\}, \nonumber \\
    f^{(1)}_{\btheta}(\bx)
    &= \sin\!\left(\omega_0 \btheta^{(1)} \bx + \bb^{(1)} + \bbeta^{(1)}\right), \nonumber \\
    f^{(\ell)}_{\btheta}(\bx)
    &= \sin\!\left(\btheta^{(\ell)} f^{(\ell-1)}_\btheta + \bb^{(\ell)} + \bbeta^{(\ell)}\right), \nonumber \\
    &\qquad\qquad\qquad\qquad\qquad \ell\in\{2,\ldots,L-1\}, \nonumber \\
    f^{(L)}_{\btheta}(\bx)
    &= \btheta^{(L)} f^{(L-1)}_\btheta + \bb^{(L)},
\end{align}
where $\btheta=\{\btheta^{(\ell)},\bb^{(\ell)}\}^{L}_{\ell=1}$, with $\btheta^{(\ell)}$ and $\bb^{(\ell)}$ denoting the weight matrix and the bias vector of the $\ell$-th layer, respectively.
Unlike hypernetworks, which generate the entire parameter set $\btheta$ for each deformation state---leading to a number of predicted parameters that grow quadratically with width of the implicit network---our approach predicts only per-channel shift parameters ${\bbeta^{(\ell)}}$, whose dimensionality scales linearly with the width of the implicit network.
This parameter-efficient conditioning strategy enables us to increase the width of the implicit network to enhance representational capacity without incurring excessive per-instance prediction costs.

The detailed network architecture is as follows.
The implicit network $f_\btheta$ is an MLP with five hidden layers of $256$ units.
Each modulation network $\calG^{(\ell)}$ is an MLP with two hidden layers, each containing $256$ units with ReLU activations.
In this work, we set the scaling factor $\omega_0=8$.

\subsection{Training for SDF Reconstruction}
\label{subsec:loss}

We introduce a learning objective that enables the implicit network $f_{\btheta}$ 
to effectively learn SDFs under diverse deformation states.
We formulate the learning objective $\calL$ as a weighted sum of
a surface supervision loss, an Eikonal regularization loss, and a latent space regularization loss:
\begin{equation}
\calL = \calL_{\text{surface}}
+ \lambda_{\mathrm{eik}} \calL_{\mathrm{eik}}
+ \lambda_{\mathrm{z}} \calL_{\mathrm{z}},
\end{equation}
where $\lambda_{\text{eik}}$ and $\lambda_{\text{z}}$ are non-negative weight coefficients.
We describe each loss term in detail below.

\begin{itemize}[leftmargin=*]
\item $\calL_{\text{surface}}$:
The surface supervision loss constrains the implicit network to represent the object surface accurately by enforcing both zero-level set consistency and correct surface orientation.
Given a complete point cloud $\calP^* = \{\bp_i\}_{i=1}^{N}$ and the corresponding surface normal vectors $\calN = \{\bn_{\bp_i}\}_{i=1}^{N}$,
where $N$ denotes the number of surface points,
we define the surface supervision loss as
\begin{align}
\label{eq:loss_surface}
\calL_{\text{surface}}
&= \frac{1}{|\calP^*|}
\sum_{\bp \in \calP^*}
\lambda_{\mathrm{surface}}
\;\mathrm{Focal}\!\left(|f_{\btheta}(\bp)|, 0\right)
\nonumber \\
&+
\lambda_{\mathrm{normal}}
\;\mathrm{Focal}\!\left(
\frac{\nabla_{\bp} f_{\btheta}(\bp)}
{\|\nabla_{\bp} f_{\btheta}(\bp)\|}
\cdot \bn_{\bp},
1
\right),
\end{align}
where $\lambda_{\text{surface}}$ and $\lambda_{\text{normal}}$
are non-negative weight coefficients.
The first term in Eq.~\eqref{eq:loss_surface} penalizes non-zero SDF values on the surface points,
while the second term encourages the normalized SDF gradients to align with the ground-truth surface normals.
To compensate for the dataset imbalance arising from the sparse occurrence of highly deformed configurations,
we adopt a focal loss formulation~\cite{lin2017focal}.
We adapt the focal loss, originally proposed for classification, to a regression setting as
\begin{equation}
\label{eq:focal}
\mathrm{Focal}(\hat{y}, y)
=
\left(\exp(|\hat{y} - y|) - 1\right)^{\alpha}
\,|\hat{y} - y|,
\end{equation}
where $\hat{y}$ denotes the predicted value,
$y$ denotes the ground-truth target value,
and $\alpha$ controls the degree of emphasis on large residuals.

\item $\calL_{\mathrm{eik}}$:
To provide supervision at off-surface query points $\bx\in\calQ$ where ground-truth signed distance values are unavailable, we employ an Eikonal regularization loss~\cite{gropp2020implicit}.
For DOs with infinite DoF, computing ground-truth signed distances at arbitrary query points
for all possible deformation states is computationally intractable.
Instead, we exploit the Eikonal equation~\cite{crandall1983viscosity},
which states that a SDF satisfies $\|\nabla_\bx f_\btheta(\bx)\| = 1$ almost everywhere.
Following IGR~\cite{gropp2020implicit},
we impose this property on off-surface query points via
\[
    \calL_{\mathrm{eik}}
    = \frac{1}{|\calQ|}
    \sum_{\bx \in \calQ}
    \left| \|\nabla_\bx f_\btheta(\bx)\| - 1 \right|.
\]

\item $\calL_{\mathrm{z}}$:
To prevent unbounded growth of the latent code,
we penalize the squared Euclidean norm of $\bz^t$,
\[
    \calL_{\mathrm{z}} = \|\bz^t\|_2^2 .
\]
This regularization stabilizes training and improves generalization to unseen deformation states.
\end{itemize}
In this work, we set $\lambda_{\mathrm{surface}}=3000, \lambda_{\mathrm{normal}}=50, \lambda_{\mathrm{eik}}=50, \lambda_{\mathrm{z}}=0.001$, and, $\alpha=2$.

%% file: sec/5_experiment.tex
\section{Experimental setup}

We evaluate the occlusion-robust reconstruction capability of ParCo-SDF on \emph{unseen deformed states} in a rubber band manipulation dataset~\cite{song2025implicit}.
The dataset consists of five rubber band deformation sequences generated by random manipulation actions in Isaac Sim, a NVIDIA physics simulator~\cite{isaacsim}.
Each sequence contains \num{2000} temporally ordered observations.
We train the model on four manipulation sequences and evaluate it on a held-out sequence with a deformation trajectory not encountered during training.
We conduct all experiments on a single rubber band with an inner diameter of \SI{10}{\centi\meter} and a cross-sectional diameter of \SI{5}{\centi\meter}.
Despite using a single type of rubber band, the elastic nature of rubber induces a wide range of continuously varying geometries during manipulation, including elongation, bending, and twisting.

\begin{figure}[t]
    \centerline{\includegraphics[width=\linewidth]{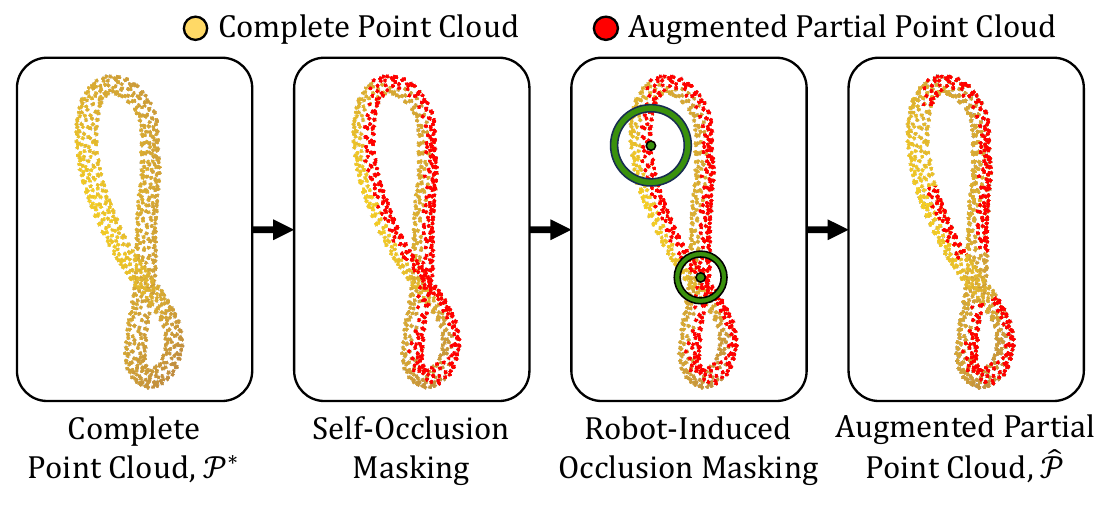}}
    \caption{
        Overview of the masked data augmentation to generate partial point-cloud observations for training.
        Starting from a complete point cloud, we select visible points based on a front-facing criterion with respect to a sampled viewing direction.
        Finally, randomly placed spherical masks remove local surface regions to emulate robot-induced occlusions.
    } \label{fig:mask_aug_process}
\end{figure}

To improve robustness to diverse occlusion patterns, we augment the training data by generating multiple partial observations from each complete point cloud.
Given a complete point cloud $\calP^*=\{\bp_i\}_{i=1}^{N}$ with corresponding surface normals $\calN=\{\bn_{\bp_i}\}_{i=1}^{N}$,
we synthesize an augmented partial point cloud $\hat{\calP}$ through a two-stage occlusion simulation process:
(i) self-occlusion masking and
(ii) robot-induced occlusion masking,
as illustrated in Fig.~\ref{fig:mask_aug_process}.
To emulate self-occlusion caused by depth sensing, we first sample a random viewing direction $\bv \sim \calU(\bbS^2)$,
where $\calU(\cdot)$ denotes a uniform distribution and $\bbS^2\subset\bbR^3$ is the unit sphere.
We regard points whose surface normals satisfy $\bn_{\bp_i}^\top \bv < 0$ as visible,
yielding the subset
$\calP_{\mathrm{vis}} = \left\{ \bp_i \;\middle|\; i \in \{1,\ldots,N\},\; \bn_{\bp_i}^\top \bv < 0\right\}$.
To further simulate occlusions caused by robot-object interactions, we remove localized surface regions from $\calP_{\mathrm{vis}}$ using multiple spherical masks.
We sample the number of masks $K \sim \calU(\{2,\ldots,6\})$.
For each mask, we sample a center $\bc_k$ uniformly from $\calP_{\mathrm{vis}}$ and a radius $r_k \sim \calU([0.15, 0.25])$, where $k \in \{1,\ldots,K\}$.
We discard points lying inside any spherical mask,
resulting in the final augmented partial point cloud:
\begin{equation}
\hat{\calP}
=
\left\{
\bp \in \calP_{\mathrm{vis}}
\;\middle|\;
\|\bp - \bc_k\|_2 > r_k,
\;\;
\forall k \in \{1,\ldots,K\}
\right\}.
\end{equation}

We introduce an evaluation metric termed the \emph{Topology Success Rate} (TSR) to quantify topological correctness under occlusion and deformation.
TSR measures the percentage of test samples whose predicted mesh has the same genus as the corresponding ground-truth mesh.
Given a predicted mesh $\hat{\calM}=(\calV,\calE,\calF)$, where $\calV$, $\calE$, and $\calF$ denote the sets of vertices, edges, and faces, respectively,
we compute the predicted genus as $\hat g = (2-\chi)/2$, with the Euler characteristic $\chi = |\calV| - |\calE| + |\calF|$.
In addition to TSR, we measure the Chamfer distance (CD) between the vertices of the reconstructed mesh and the ground-truth complete point cloud, 
thereby providing a comprehensive assessment of both topological correctness and geometric fidelity.
We extract meshes from the predicted SDFs using the marching cubes algorithm~\cite{lorensen1987marching}.

We compare our method against a state-of-the-art DO surface reconstruction baseline, INR-DOM~\cite{song2025implicit}.
INR-DOM introduces a skeleton loss that encourages the reconstruction to align with a predefined medial axis of the rubber band and conditions the implicit network through a hypernetwork-based parameterization.

\section{Evaluation Results}
\begin{figure*}[t]
    \centerline{\includegraphics[width=\linewidth]{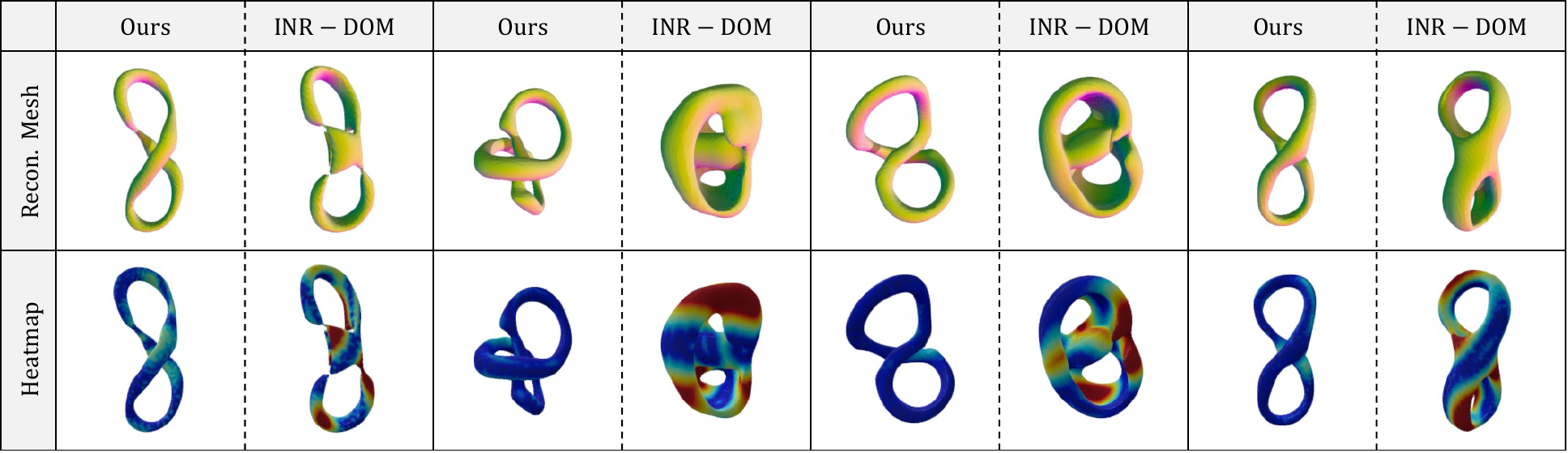}}
    \caption{
        Qualitative comparison of occlusion-robust reconstruction between ParCo-SDF and INR-DOM across four observation cases.
        (Top rows) Reconstructed meshes produced by ParCo-SDF and INR-DOM.
        (Bottom rows) Chamfer distance heatmaps on the reconstructed surfaces, where red and blue denote high and low reconstruction errors, respectively.
    } \label{fig:oring_qual_res}
\end{figure*}

\begin{figure}[t]
    \centerline{\includegraphics[width=\linewidth]{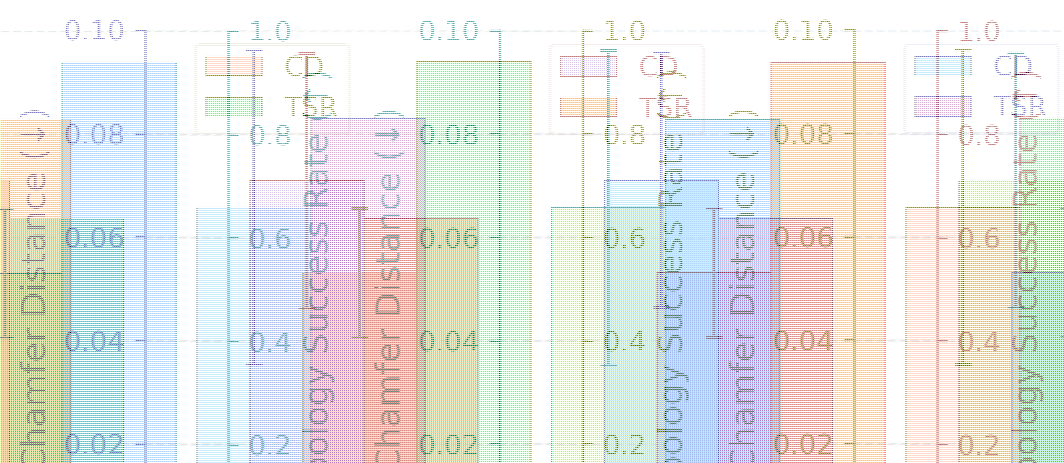}}
    \caption{
        Comparison of surface reconstruction performance for unseen deformations of elastic bands. The blue and red bars represent the reconstruction errors measured by chamfer distance (CD) and topology success rate (TSR), respectively. Note that ParCo-SDF$^{-T}$ refers to a variant of ParCo-SDF trained without temporal aggregation layers. 
    } \label{fig:oring_quan_res}
\end{figure}

We quantitatively evaluate reconstruction performance on unseen deformed states of elastic bands using two complementary metrics:
CD for geometric fidelity and TSR for topological preservation.

Fig.~\ref{fig:oring_quan_res} shows that ParCo-SDF achieves superior partial-to-complete reconstruction performance, yielding the lowest CD and the highest TSR among all compared approaches.
Compared to INR-DOM, our method reduces CD by \SI{23.6}{\percent} while improving TSR by \SI{21.9}{\percent}, demonstrating superior reconstruction accuracy and robustness to severe occlusions.
Notably, whereas INR-DOM relies on object symmetry priors, our approach consistently reconstructs the correct topology without such assumptions.
We further compare our full model with ParCo-SDF$^{-T}$, which excludes temporal aggregation layers.
The performance gap indicates that temporal aggregation plays a crucial role in inferring complete topology from sequential partial observations.

In addition, we analyze the standard deviation of reconstruction errors.
The hypernetwork-based conditioning baseline, INR-DOM, exhibits a relatively large standard deviation in CD, suggesting unstable reconstruction under varying occlusion patterns.
In contrast, both ParCo-SDF and ParCo-SDF$^{-T}$, which employ FiLM-based conditioning,
show significantly lower variance, indicating more stable and consistent reconstructions.



Fig.~\ref{fig:oring_qual_res} presents qualitative comparisons between ParCo-SDF and INR-DOM.
Despite large missing regions in the input partial observations, ParCo-SDF successfully reconstructs geometrically coherent surfaces that preserve the torus topology of the rubber band.
In contrast, INR-DOM often produces spurious connections under severe occlusions, leading to incorrect genus and surface artifacts.
Overall, the qualitative results demonstrate that leveraging temporally aggregated geometric cues significantly improve robustness to occlusion and enables accurate topology preservation.

%% file: sec/7_dicussion.tex
\section{Conclusion}
We proposed ParCo-SDF, a prior-free partial-to-complete signed distance field reconstruction method for deformable objects. By aggregating a sliding window of partial observations, our method robustly represents complete geometry without relying on explicit shape priors. Furthermore, FiLM-based conditioning of a shared implicit SDF network enables stable modeling of continuously deforming geometries. Quantitative and qualitative evaluations on rubber band manipulation scenarios show that our method significantly improves geometric fidelity and topological consistency under severe occlusions and previously unseen deformation states, outperforming the state-of-the-art baseline method.
Although our experiments were conducted in simulation and focused on a single deformable object category, the proposed framework provides a promising foundation for prior-free deformable geometry reconstruction. Future work will investigate generalization across diverse deformable objects and real-world robotic manipulation scenarios.
